\documentclass[11pt,a4paper]{article}
\usepackage[hyperref]{acl2018}
\usepackage{times}
\usepackage{latexsym}
\usepackage{float}
\usepackage{url}
\usepackage{multirow}
\usepackage{graphicx}
\usepackage{amsmath, amsfonts}
\usepackage{caption}
\usepackage{tabularx}

\aclfinalcopy 

\newcommand\Mark[1]{\textsuperscript#1}

\title{SMT vs NMT: A Comparison over Hindi \& Bengali Simple Sentences}

\author{Sainik Kumar Mahata\Mark{1},
Soumil Mandal\Mark{2}, Dipankar Das\Mark{3}, Sivaji Bandyopadhyay\Mark{4}\\
\Mark{1}\Mark{3}\Mark{4}Jadavpur University, Kolkata\\
\Mark{2}SRM University, Chennai\\
\Mark{1}sainik.mahata@gmail.com,
\Mark{2}soumil.mandal@gmail.com,\\
\Mark{3}dipankar.dipnil2005@gmail.com,
\Mark{4}sivaji\_cse\_ju@yahoo.com}

\date{}

\begin{document}
\maketitle
\begin{abstract}
In the present article, we identified the qualitative differences between Statistical Machine Translation (SMT) and Neural Machine Translation (NMT) outputs. We have tried to answer two important questions: \textbf{1.} Does NMT perform equivalently well with respect to SMT and \textbf{2.} Does it add extra flavor in improving the quality of MT output by employing simple sentences as training units. In order to obtain insights, we have developed three core models viz., SMT model based on Moses toolkit, followed by character and word level NMT models. All of the systems use English-Hindi and English-Bengali language pairs containing simple sentences as well as sentences of other complexity. In order to preserve the translations semantics with respect to the target words of a sentence, we have employed soft-attention into our word level NMT model. We have further evaluated all the systems with respect to the scenarios where they succeed and fail. Finally, the quality of translation has been validated using BLEU and TER metrics along with manual parameters like fluency, adequacy etc. We observed that NMT outperforms SMT in case of simple sentences whereas SMT outperforms in case of all types of sentence.

\end{abstract}

\section{Introduction}
Machine Translation (MT) refers to automated translation. It is the process by which computer software is used to translate a text from one natural language (such as English) into another (such as Spanish). Translation itself is a challenging task for humans, and hence, is more challenging for computers. High quality translation requires a thorough understanding of syntax and semantical properties of both the source and target languages.

The importance of studying and developing better MT systems has gained popularity in the recent past due to rapid globalization, where people from multiple backgrounds having variety of language knowledge are working together. Primarily, two paradigms are currently followed for building MT systems. One is based on statistical techniques, while the other employs artificial neural networks.

The statistical model, commonly referred to as Statistical Machine Translation (SMT) \cite{weaver:1955translation}, addresses this challenge by creating statistical models, where input parameters are derived from the analysis of parallel bilingual text corpora \cite{mahata:2017bucc2017}. Some of the notable works on SMT are \cite{al:1999statistical,lopez:2008statistical,koehn:2009statistical}, where the authors have dived deep into various challenges, working principles and possible improvements. SMT has shown good results for many language pairs and is responsible for the recent surge in the popularity of MT  among general public \nocite{mahataMTIL}. 

On the other hand, despite being relatively new, Neural Machine Translation (NMT) \cite{bahdanau:2014neural} has already shown promising results \cite{mahata:2016wmt2016, wu:2016google} and hence has gained substantial attention as well as interest. Continuous recurrent models for translation, which do not depend on alignment or phrasal translation units, was introduced by \citet{kalchbrenner:2013recurrent}. On the other hand, the problem of rare word occurrence was addressed by \citet{luong:2014addressing} and the effectiveness of global and local approach was explored by \citet{luong:2015effective}. \citet{he:2016improved} demonstrated a log-linear framework by incorporating SMT features combined with NMT which addresses the issues like out of vocabulary and inadequate translation. The properties of these architecture were discussed in detail in \cite{cho:2014properties}. This approach generally produces much more accurate translation than SMT even with the adequate supply of training data. \cite{vaswani:2013decoding,liu:2014recursive,doherty:2010eye}.

In the current work, we have tested the performance of SMT and NMT on simple sentences (see Section \ref{simplesentence}) extracted from English-Hindi (En-Hn) and English-Bengali (En-Bn) parallel corpora provided by TDIL\footnote{http://www.tdil.meity.gov.in/}. These experiments were done to dive into the scenarios where NMT and SMT outperform each other. Moreover, they would also help us in evaluating the question that whether usage of simple sentences as training data for MT models really evokes any difference in the quality of the MT output or not.  

We have constrained our target language domain to Hindi and Bengali as these languages are used primarily in the Indian sub-continent. Number of native speakers of Hindi in India is 41.1\% while that of Bengali is 8.11\%. Hindi is written in Devanagari~\footnote{https://en.wikipedia.org/wiki/Devanagari} script and Bengali is written in Eastern Nagari~\footnote{https://en.wikipedia.org/wiki/Eastern\_Nagari\_script} script. 

In order to test the effectiveness of the case study, SMT and NMT systems were also trained for the whole corpus which consists of sentences with mixed complexity. For both simple sentence corpus and the whole corpus, BLEU \cite{Papineni02bleu:a}, TER \cite{Snover:06astudy} and manual evaluation metrics like fluency and adequecy were calculated to validate the observed results.

The paper has been organized as follows. Section \ref{simplesentence} describes the extraction of simple sentences from the parallel corpus given by TDIL. Section \ref{SMT} and Section \ref{NMT}, describes the methodology for the training of the SMT and the NMT models, respectively. Later, Section \ref{evaluation} describes the evaluation with respect to various metrics and finally, Section \ref{conclusion} draws the conclusion.

\section{Extraction of Simple Sentence}
\label{simplesentence}
Since we wanted to analyze and compare both the models viz. SMT and NMT with respect to how they perform on simple sentences, we first needed to extract such instances from our dataset that had data of varying complexity. 

A simple sentence in this context is defined as a sentence that contains only one independent clause and has no dependent clauses. Generally, whenever two or more clauses are joined by conjunctions (coordinating and subordinating), it becomes a complex or a compound sentence accordingly. So, to get a hold on handling the conjunctions, we used the Stanford Dependency Parser~\footnote{https://stanfordnlp.github.io/CoreNLP/} library to chunk the English sentences into phrases. (viz. NP (Noun Phrase), VP (Verb Phrase), PP (Preposition Phrase), ADJP (Adjective Phrase) and ADVP (Adverb Phrase)). 
\begin{figure}[h]
\centering
\includegraphics[scale=0.4]{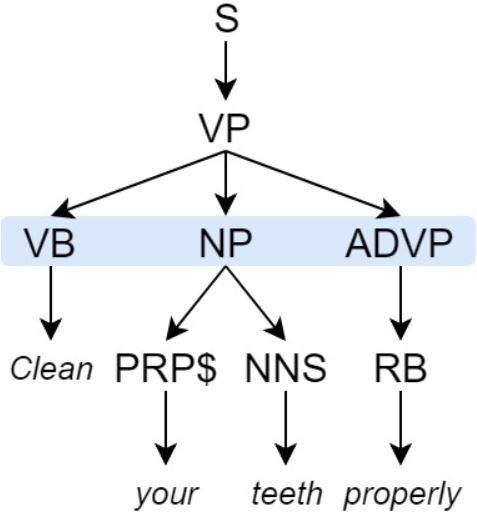}
\caption{Extraction of phrase chunks.}
\label{fig_1}
\end{figure}

We noticed that, simple sentences have an unique phrase structure that consists of some combinations of NP, VP and PP. In conjunction with this theory, we applied two approaches (viz. rule based approach and deep learning based approach) to extract simple sentences from the English corpus. The approaches are discussed in Section \ref{RB Simple} and Section \ref{DL Simple}, respectively.

\subsection{Rule based Approach}
\label{RB Simple}
We subjected a total of 3046 simple sentences, extracted from various websites, to chunk using Stanford Dependency Parser \cite{manning:2014stanford}, and identified their unique phrase structures. Such structures became the rules by which we further mined simple sentences from the English corpus. 

We  extracted 205 unique rules, the surface forms of which, along with its Confidence Score, are shown in Table \ref{surface forms}. The confidence score of the rules were calculated using

\small
$$Confidence Score = \frac{\# sentences pertaining to a rule}{ \#  total sentences in the test data}$$
\normalsize

We tested our system on 2876 sentences (1438 simple sentences and 1438 complex/compound sentences) and achieved an accuracy of 89.22\%. Table \ref{accuracy matrix1} shows the various validation metrics. Using this system, 10,349 simple sentences from the TDIL English corpus were extracted, as shown in Table \ref{simple count}. 
\begin{table}[h]
\begin{tabular}{|l|c|}
\hline
\textbf{Rules} & \textbf{Confidence} \\ \hline
PP NP* PP VP NP* & 8.40 \\ \hline
PP NP* VP PP NP* & 9.49 \\ \hline
ADVP NP* VP* ADVP NP* & 9.36  \\ \hline
NP VP PP NP PP NP & 12.15 \\ \hline
NP ADVP VP* NP* & 11.69 \\ \hline
NP* VP NP* & 11.69 \\ \hline
NP* PP NP VP* NP & 11.46 \\ \hline
NP VP PP NP* & 11.23 \\ \hline
VP* NP* PRP* ADVP* & 4.92 \\ \hline
NP VP* NP* PP* ADJP* ADVP* & 9.62 \\ \hline
\end{tabular}
\captionsetup{justification=centering}
\caption{Surface forms of the extracted rules. "*" means one or more occurrence of item.}
\label{surface forms}
\end{table}

\begin{table}[h]
\centering
\scalebox{0.9}{
\begin{tabular}{|l|c|c|c|c|}
\hline
\multicolumn{1}{|c|}{\textbf{}} & \textbf{Other} & \textbf{Simple} & \textbf{Prec.} & \textbf{Kappa} \\ \hline
\textbf{Other} & 1275 & 90 & \multirow{2}{*}{93.41\%} & \multirow{3}{*}{0.78} \\ \cline{1-3}
\textbf{Simple} & 220 & 1291 &  &  \\ \cline{1-4}
\textbf{Recall} & \multicolumn{2}{c|}{85.28\%} &  &  \\ \hline
\textbf{Acc.} & \multicolumn{4}{c|}{89.22\%} \\ \hline
\textbf{F1} & \multicolumn{4}{c|}{89.16\%} \\ \hline
\end{tabular}}
\captionsetup{justification=centering}
\caption{Confusion matrix for the rule based approach.}
\label{accuracy matrix1}
\end{table}
 
\subsection{Deep Learning based Approach}
\label{DL Simple}
We preferred Deep Learning approach over traditional Machine Learning (ML) approach as because in the ML approach we could only extract syntactic features, which was already exploited in the rule based approach discussed in Section \ref{RB Simple}. On the other hand, a deep learning technique learn categories incrementally through it’s hidden layer architecture. We wanted the deep learning framework to learn the nature of a sentence from the POS tags itself as it automatically clusters similar data into separate spaces.

For the deep learning model, we trained a multi-layer feed-forward neural network with stochastic gradient descent \cite{bottou:2010large} as optimizer with back-propagation. The network contained two hidden layers of sizes 50 and 50, respectively. The activation function used was \textit{tanh} and loss function used was \textit{Mean Squared Error}. \textit{Learning Rate} was kept at 0.001 and \textit{number of epochs} were fixed at 100. The \textit{batch size} was kept at 128. The training data consisted phrases of 2876 sentences (1438 simple sentences and 1438 other complex/compound sentences). The trained model was subjected to 10 fold cross validation and it yielded an accuracy figure of 92.11\%. Table~\ref{accuracy matrix} shows the results with respect to other important validation metrics.
\begin{table}[h]
\centering
\scalebox{0.9}{
\begin{tabular}{|l|c|c|c|c|}
\hline
\multicolumn{1}{|c|}{\textbf{}} & \textbf{Other} & \textbf{Simple} & \textbf{Prec.} & \textbf{Kappa} \\ \hline
\textbf{Other} & 1287 & 76 & \multirow{2}{*}{92.22\%} & \multirow{3}{*}{0.84} \\ \cline{1-3}
\textbf{Simple} & 151 & 1362 &  &  \\ \cline{1-4}
\textbf{Recall} & \multicolumn{2}{c|}{92.11\%} &  &  \\ \hline
\textbf{Acc.} & \multicolumn{4}{c|}{92.11\%} \\ \hline
\textbf{F1} & \multicolumn{4}{c|}{92.16\%} \\ \hline
\end{tabular}}
\captionsetup{justification=centering}
\caption{Confusion matrix for deep learning based approach.}
\label{accuracy matrix}
\end{table}

\noindent The TDIL English corpus was fed to this model and it yielded 14,976 simple sentences as shown in Table \ref{simple count}. 

\begin{table}[h]
\centering
\scalebox{0.9}{
\begin{tabular}{|l|c|c|}
\hline
\textbf{\# of sentences} & \textbf{} & 49999 \\ \hline
\textbf{\# of other sentences} & RL & 39650 \\ \hline
\textbf{\# of simple sentences} & RL & 10349 \\ \hline
\textbf{\# of other sentences} & DL & 35023 \\ \hline
\textbf{\# of simple sentences} & DL & 14976 \\ \hline
\end{tabular}}
\caption{Simple Sentence Count}
\label{simple count}
\end{table}

The deep learning based approach was preferred as it resulted better accuracy. The Bengali and Hindi counterparts of these sentences were extracted to build a parallel corpus comprising of simple sentences only. The next step was to build MT models using this data, as well as the data from the whole corpus, and compare their respective results.

\section{Statistical Machine Translation}
\label{SMT}
We know that Moses \cite{koehn:2007moses} is a statistical machine translation system that allows us to automatically train translation models for any language pair, making use of a large collection of translated texts (parallel corpus). Once the model has been trained, an efficient beam search algorithm quickly finds the highest probability translation among the exponential number of choices. 

For training the SMT model, we used English as the source language and Bengali and Hindi as the target languages. To prepare the data for training the SMT system, we performed the following steps.

\subsection{Preprocessing}
The following steps were employed to preprocess the Source and the Target texts.
\begin{itemize}
\itemsep 0em 
\item \textbf{Tokenization}: Given a character sequence and a defined document unit, tokenization is applied for chopping it up into pieces, called tokens. In our case, these tokens were words, punctuation marks, numbers.
\item \textbf{Truecasing}: This refers to the process of restoring case information to badly-cased or non-cased text \cite{lita:2003truecasing}. Truecasing helps in reducing data sparsity.
\item \textbf{Cleaning}: Long sentences (\# of tokens $>$ 80) were removed.
\end{itemize}

\subsection{Language Model}
A \textit{Language Model} (LM) was built using the target language, Bengali and Hindi, in our case, to ensure fluent output. KenLM \cite{heafield:2011kenlm}, which comes bundled with the Moses toolkit, was used for building this model.

\subsection{Word Alignment and Phrase Table Generation}
For word alignment in the translation model, GIZA++ \cite{och03:asc} was used. Finally, the phrase table was created and probability scores were calculated. Training the Moses statistical MT system resulted in the generation of two models, one is a Phrase Model and the other is a Translation Model. Moses scores the phrase in the phrase table with respect to a given source sentence and produces best scored phrases as output. 

The results and evaluation of this system are shown in Sec~\ref{evaluation}, Table~\ref{Table5} and Table~\ref{Table6} when trained and tested on the simple sentence corpus and the general corpus, for both En-Bn and En-Hn language pairs. 

\section{Neural Machine Translation}
\label{NMT}
Neural machine translation (NMT) is a MT approach that uses neural networks to predict the likelihood of a sequence of words, typically modeling entire sentences in a single integrated model. NMT departs from traditional phrase-based statistical approaches in the sense that it uses separately engineered subcomponents like Language Model generation, Word Alignment and Phrase Table generation. The main functionality of NMT is based on the sequence to sequence (seq2seq) architecture, which is described in Section \ref{seq2seq}.
\begin{figure*}[h]
\fbox{\includegraphics[height=6cm, width=\textwidth]{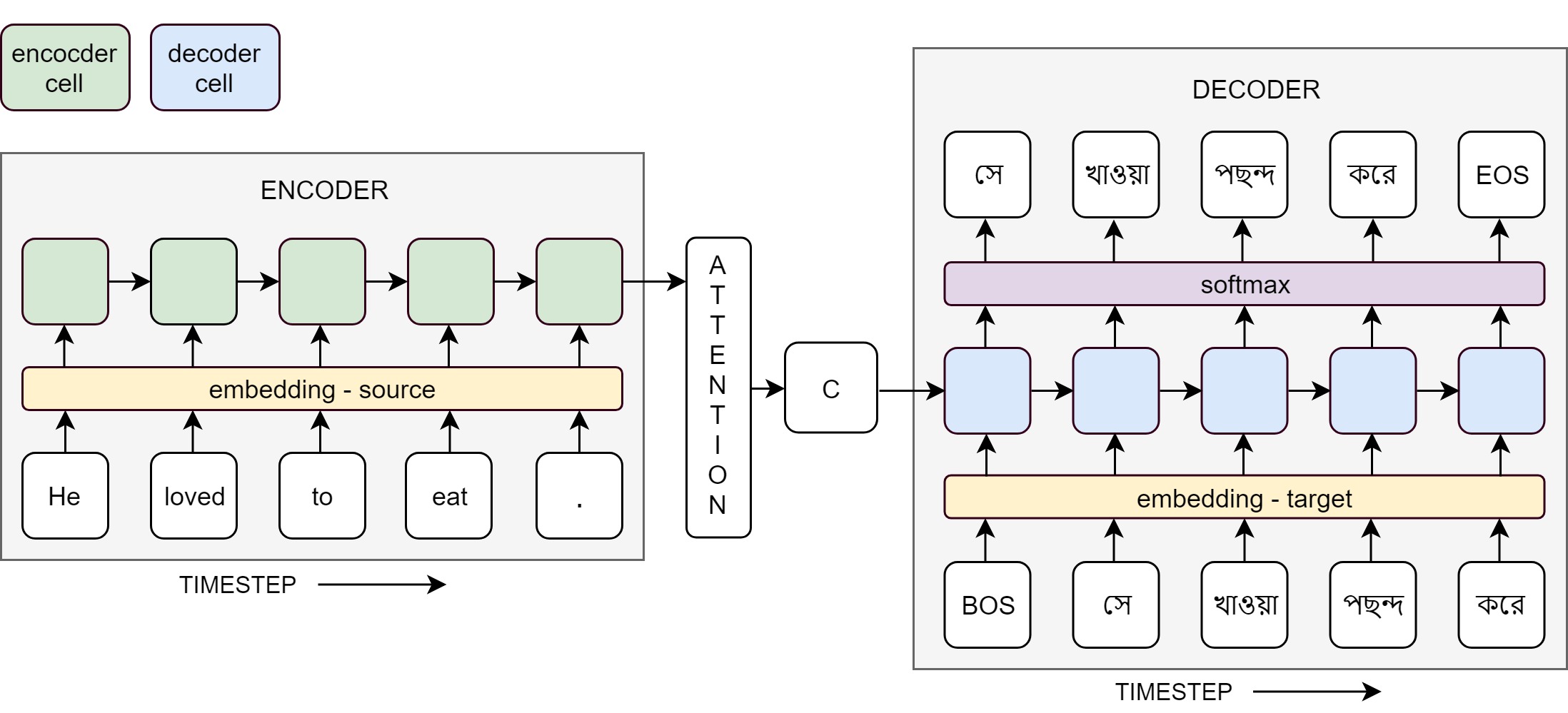}}
\caption{NMT with attention architecture.}
\label{fig_2}
\end{figure*}
\subsection{Seq2Seq Model}
\label{seq2seq}
The sequence to sequence model is a relatively new idea for sequence learning using neural networks. It has gained quite some popularity since it achieved state of the art results in machine translation task. Essentially, the model takes a sequence as input  
\begin{equation}
\notag
X=\{x\textsubscript{1}, x\textsubscript{2}, ..., x\textsubscript{n}\} 
\end{equation}
and tries to generate the target sequence as output
\begin{equation}
\notag
Y = \{y\textsubscript{1}, y\textsubscript{2}, ..., y\textsubscript{m}\}
\end{equation}
where x\textsubscript{i} and y\textsubscript{i} are the input and target symbols, respectively. The architecture of seq2seq model comprises of two parts, the encoder and decoder. We experimented with two types of NMT models (word and character level) and both the models use the seq2seq architecture, the difference being in the inputs to its encoder and decoder. They are discussed in the sections \ref{SMT} and \ref{NMT} below. The working architecture of seq2seq model at the word level is shown in Fig. \ref{fig_2}. We implemented both the models using the Keras \cite{chollet:2015keras} library.

\subsubsection{Word Level NMT}
To build our world level NMT model, we used the seq2seq with attention mechanism. This architecture has recently shown to achieve state of the art quality translation across many different language pairs. The details of the seq2seq model along with the training details are given below.

\paragraph{Encoder}
The encoder takes a variable length sequence as input and encodes it into a fixed length vector, which is supposed to summarize it's meaning and taking into account it's context as well. A Long Short Term Memory (LSTM) cell was used to achieve this. The directional encoder reads the sequence from one end to the other (left to right in our case), 
\begin{equation}
\notag
\vec{h}\textsubscript{t} = \vec{f}\textsubscript{enc}(E\textsubscript{x}(x\textsubscript{t}),\vec{h}\textsubscript{t-1})
\end{equation}
Here, E\textsubscript{x} is the input embedding lookup table (dictionary), $\vec{f}$\textsubscript{enc} are the transfer function for the Long Short Term Memory (LSTM) recurrent unit \cite{hochreiter:1997long}. A contiguous sequence of encodings C is constructed and then passed on to the decoder.

\paragraph{Decoder} 
The decoder takes as input, the context vector C from the encoder, and computes the hidden state at time t as, 
\begin{equation}
\notag
s\textsubscript{t} =  f\textsubscript{dec}(E\textsubscript{y}(y\textsubscript{t-1}), s\textsubscript{t-1}, c\textsubscript{t})
\end{equation} 
Subsequently, a parametric function out\textsubscript{k} returns the conditional probability using the next target symbol k. \\
\begin{equation}
\notag
(y\textsubscript{t}=k\mid y<{t}, X) = \frac{1}{Z}exp(out\textsubscript{k}(E\textsubscript{y}(y\textsubscript{t}-1), s\textsubscript{t}, c\textsubscript{t}))
\end{equation}
Z is  the normalizing constant, 
\begin{equation}
\notag
\sum\textsubscript{j}exp(out\textsubscript{j}(E\textsubscript{y}(y\textsubscript{t}-1), s\textsubscript{t}, c\textsubscript{t}))
\end{equation} 
The entire model can be trained end-to-end by minimizing the log likelihood which is defined as
\begin{equation}
\notag
L = -\frac{1}{N}\sum_{n=1}^{N}\sum_{t=1}^{T\textsubscript{y}\textsuperscript{n}}log p(y\textsubscript{t} = y\textsubscript{t}\textsuperscript{n}, y\textsubscript{<t}\textsuperscript{n}, X\textsuperscript{n})
\end{equation}
where N is the number of sentence pairs, and X\textsuperscript{n} and y\textsubscript{t}\textsuperscript{n} are the input sentence and the t-th target symbol in the n\textsuperscript{th} pair, respectively.  

\paragraph{Training} 
For training our model, we used the seq2seq with attention architecture by employing LSTM cell. We used two LSTM cells, stacked upon each other, where one acts as the encoder and the other as the decoder. We trained our model on 14976 data (for simple sentence corpus), 49999 sentences (for Bengali and Hindi whole general corpus), \textit{batch size} at 256, \textit{number of epochs} at 100 and \textit{learning rate} at 0.001. The activation function used was \textit{softmax}, optimizer used was \textit{rmsprop} and the loss calculation at each step was done using \textit{categorical cross-entropy}. 

\paragraph{Attention}
\label{attention}
Neural processes involving attention \cite{vaswani:2017attention} has been largely studied in computational neuro-science. This concept is very loosely based on visual attention mechanism in humans. With attention mechanism, the need to encode the full source sentence into a fixed length vector is omitted. Rather we allow the decoder to attend different parts of the source sentence at each time step of the output generation. Essentially, we let the model learn what to \textit{attend} based on the input sequence and what is predicted so far. 

Mathematically, it computes the context vector c\textsubscript{t} at each time step t as a weighted sum of the source hidden states, 
\begin{equation}
\notag
c\textsubscript{t} = \sum\textsubscript{t=1}\textsuperscript{Tx}\alpha\textsubscript{t}h\textsubscript{t}
\end{equation} 
Each attention weight $\alpha$\textsubscript{t} represents how much relevant the t\textsuperscript{th} source token x\textsubscript{t} is to the t\textsuperscript{th} target token y\textsubscript{t} and is computed as :
\begin{equation}
\notag
\alpha\textsubscript{t} = \frac{1}{Z}exp(score(E\textsubscript{y}(y\textsubscript{t}-1), s\textsubscript{t-1}, h\textsubscript{t}))
\end{equation}
where 
\begin{equation}
\notag
Z = \sum\textsubscript{k=1}\textsuperscript{Tx}exp(score(E\textsubscript{y}(y\textsubscript{t}-1), s\textsubscript{t-1}, h\textsubscript{k}))
\end{equation}
Z is the normalization constant. score() is a feed forward neural network with a single hidden layer that scores how well the source symbol x\textsubscript{x} and the target symbol y\textsubscript{t} match. E\textsubscript{y} is the target embedding lookup table and s\textsubscript{t} is the target hidden state at time t. The results and evaluation of the systems are shown in Section~\ref{evaluation}.

\subsubsection{Character Level NMT}
\label{CNMT}
It was observed that Character level NMT (CNMT) performs better than Word level NMT (WNMT) due to the following reasons \cite{DBLP:journals/corr/ChungCB16}
\begin{enumerate}
\setlength{\itemsep}{0pt}
\item It does not suffer from out-of-vocabulary issues
\item It is able to model different, rare morphological
variants of a word
\item It does not require segmentation. 
\end{enumerate}
Generally, CNMT works the best when majority of alphabets, in the source and target language, overlap i.e both the languages share a common or similar script. Still, we tried to find out its performance on the simple sentence and whole corpus, though in our case, Nagari script and Roman script utilizes completely different alphabets. The model has two parts (encoder and decoder) as discussed below.

\paragraph{Encoder}
In order to build the encoder, we used LSTM cells. The input of the cell was one hot tensor of English sentences (embeddings at character level). From the encoder, the internal states of each cell were preserved and the outputs were discarded. The purpose of this is to preserve the information at context level. These states were then passed on to the decoder cell as initial states. 

\paragraph{Decoder} 
However, for building the decoder, again an LSTM cell was used with initial states as the hidden states from encoder. It was designed to return both sequences and states. The input to the decoder was one hot tensor (embeddings at character level) of Bengali and Hindi sentences while the target data was identical, but with an offset of one time-step ahead.  The information for generation is gathered from the initial states passed on by the encoder. Thus, the decoder learns to generate target data [t+1,...] given targets [..., t] conditioned on the input sequence. It essentially predicts the output sequence, one character per time step.

\paragraph{Training}
For training the model, \textit{batch size} was set to 64, \textit{number of epochs} was set to 100, activation function was \textit{softmax}, optimizer chosen was \textit{rmsprop} and loss function used was \textit{categorical cross-entropy}. Learning rate was set to 0.001. The results and evaluation of the systems are shown in Section~\ref{evaluation}.

\section{Evaluation and Analysis}
All of our translation systems were evaluated in two ways, automatic and manual, depictions of which are discussed in the section below.
\label{evaluation}
\subsection{Automatic Evaluation}
Automatic evaluation was done by scoring the translations using BLEU and TER metrics. The results are shown in Table \ref{Table5} and \ref{Table6}. In the tables, "Bn" and "Hn" means Bengali and Hindi, respectively. "CNMT" and "WNMT" means character and word level NMT models, respectively. The presence of attention mechanism in the model is denoted using "A" and the contrary is denoted using "NA"

\begin{table}[h]
\centering
\scalebox{0.9}{
\begin{tabular}{|l|l|c|c|c|c|}
\hline
\multicolumn{2}{|c|}{\multirow{2}{*}{\textbf{\begin{tabular}[c]{@{}c@{}}Model \\ (Bn)\end{tabular}}}} & \multicolumn{2}{c|}{\textbf{Simple Sent.}} & \multicolumn{2}{c|}{\textbf{Whole Corp.}} \\ \cline{3-6} 
\multicolumn{2}{|c|}{} & \textbf{BLEU} & \textbf{TER} & \textbf{BLEU} & \textbf{TER} \\ \hline
\multicolumn{2}{|l|}{\textbf{SMT}} & 0 & 117.67 & 15.9 & 85.26 \\ \hline
\multicolumn{2}{|l|}{\textbf{CNMT (NA)}} & 8.69 & 91.87 & 4.19 & 88.22 \\ \hline
\multicolumn{2}{|l|}{\textbf{WNMT (NA)}} & 9.68 & 86.84 & 3.61 & 98.03 \\ \hline
\multicolumn{2}{|l|}{\textbf{WNMT (A)}} & 9.95 & 85.66 & 3.77 & 96.72 \\ \hline
\end{tabular}}
\captionsetup{justification=centering}
\caption{Automatic evaluation metrics for En-Bn Model.}
\label{Table5}
\end{table}

\begin{table}[h]
\centering
\scalebox{0.9}{
\begin{tabular}{|l|l|c|c|c|c|}
\hline
\multicolumn{2}{|c|}{\multirow{2}{*}{\textbf{\begin{tabular}[c]{@{}c@{}}Model \\ (Hn)\end{tabular}}}} & \multicolumn{2}{c|}{\textbf{Simple Sent.}} & \multicolumn{2}{c|}{\textbf{Whole Corp.}} \\ \cline{3-6} 
\multicolumn{2}{|c|}{} & \textbf{BLEU} & \textbf{TER} & \textbf{BLEU} & \textbf{TER} \\ \hline
\multicolumn{2}{|l|}{\textbf{SMT}} & 3.98 & 101.945 & 12.86 & 95.092 \\ \hline
\multicolumn{2}{|l|}{\textbf{CNMT (NA)}} & 7.98 & 92.85 & 5.96 & 85.18 \\ \hline
\multicolumn{2}{|l|}{\textbf{WNMT (NA)}} & 10.01 & 90.28 & 4.87 & 96.97 \\ \hline
\multicolumn{2}{|l|}{\textbf{WNMT (A)}} & 10.54 & 90.26 & 5.21 & 94.20 \\ \hline
\end{tabular}}
\captionsetup{justification=centering}
\caption{Automatic evaluation metrics for En-Hn Model.}
\label{Table6}
\end{table}

\subsection{Manual Evaluation}
\begin{table*}[ht]
\centering
\scalebox{0.9}{
\begin{tabular}{|c|c|c|c|c|c|l|c|l|}
\hline
\textbf{Model (Bn)}    & \multicolumn{2}{c|}{\textbf{SMT}} & \multicolumn{2}{c|}{\textbf{CNMT}} & \multicolumn{2}{c|}{\textbf{WNMT (NA)}} & \multicolumn{2}{c|}{\textbf{WNMT(A)}} \\ \hline
\textbf{Corpus}        & \textbf{Simple}  & \textbf{Whole} & \textbf{Simple}  & \textbf{Whole}  & \textbf{Simple}     & \textbf{Whole}    & \textbf{Simple}    & \textbf{Whole}   \\ \hline
\textbf{Adequecy 1}    & 0            & 2.15            & 1.98             & 1.54            & 2.02                 & 1.44              & 2.15               & 1.47             \\ \hline
\textbf{Fluency 1}     & 0             & 1.87           & 2.27             & 1.98            & 2.36               & 1.86              & 1.98               & 2.02             \\ \hline
\textbf{Adequecy 2}    & 0             & 2.24           & 1.87             & 1.66            & 1.96                & 1.57              & 2.01               & 1.69             \\ \hline
\textbf{Fluency 2}     & 0             & 1.92           & 2.05             & 1.86            & 2.21                & 1.77              & 2.26               & 1.93             \\ \hline
\textbf{Avg. Adequecy} & 0            & 2.195          & 1.925            & 1.6           & 1.99                & 1.505             & 2.08               & 1.58             \\ \hline
\textbf{Avg. Fluency}  & 0            & 1.895          & 2.16             & 1.92           & 2.285                & 1.815             & 2.12              & 1.975             \\ \hline
\end{tabular}}
\captionsetup{justification=centering}
\caption{Depiction of Manual Evaluation conducted by Bengali language speaking experts.}
\label{Table7}
\end{table*}

\begin{table*}[ht]
\centering
\scalebox{0.9}{
\begin{tabular}{|c|c|c|c|c|c|c|c|c|}
\hline
\textbf{Model (Hn)} & \multicolumn{2}{c|}{\textbf{SMT}} & \multicolumn{2}{c|}{\textbf{CNMT}} & \multicolumn{2}{c|}{\textbf{WNMT(NA)}} & \multicolumn{2}{c|}{\textbf{WNMT(A)}} \\ \hline
\textbf{Corpus} & \textbf{Simple} & \textbf{Whole} & \textbf{Simple} & \textbf{Whole} & \textbf{Simple} & \textbf{Whole} & \textbf{Simple} & \textbf{Whole} \\ \hline
\textbf{Adequecy 1} & 0.8 & 2.06 & 1.96 & 1.69 & 2.36 & 1.47 & 2.26 & 1.49 \\ \hline
\textbf{Fluency 1} & 0.5 & 1.72 & 2.04 & 2.08 & 2.27 & 1.92 & 2 & 2.22 \\ \hline
\textbf{Adequecy 2} & 1.02 & 2.18 & 1.79 & 1.71 & 2.02 & 1.63 & 2.18 & 1.9 \\ \hline
\textbf{Fluency 2} & 0.65 & 1.98 & 2.1 & 1.94 & 2.39 & 1.83 & 2.33 & 1.87 \\ \hline
\textbf{Avg. Adequecy} & 0.91 & 2.12 & 1.875 & 1.7 & 2.19 & 1.55 & 2.22 & 1.695 \\ \hline
\textbf{Avg. Fluency} & 0.575 & 1.85 & 2.07 & 2.01 & 2.33 & 1.875 & 2.165 & 2.045 \\ \hline
\end{tabular}}
\captionsetup{justification=centering}
\caption{Depiction of Manual Evaluation conducted by Hindi language speaking experts. }
\label{Table8}
\end{table*}

Translation quality was judged by four linguists. Two had Bengali mother tongue (evaluated Bn model), while the other two had Hindi mother tongue (evaluated Hn model). The evaluation criteria were Adequacy and Fluency. Adequacy means how much of the meaning expressed in the target translation. Fluency means to what extent the translation is well-formed grammatically, contains correct spellings and intuitively acceptable and can be sensibly interpreted by a native speaker. The speakers were asked to rate the translation in range of 1-5, where '1' is the lowest and '5' is the highest. The manual evaluation measures for English-Bengali and English-Hindi language pair are given in Table \ref{Table7} and Table \ref{Table8}, respectively. 

\subsection{Analysis}

We can clearly see in the results, that a NMT model, when trained using simple sentences, performs better than a SMT model, when trained using the same sentence pairs. 

But, at the same time, SMT outperforms NMT, when trained using the whole corpus. This is due to the fact that NMT doesn't quite work well with less amount of data and highly complex sentences. 

Similarly, we also see that character based NMT works better than word based NMT, when dealing with less amount of data. But again, we have to keep in mind that for a character based NMT to work well, we have to train it using a Source-Target language pair, who share a common script. 

Further, word based NMT with attention perform relatively better than a character based NMT. We didn't use attention in the Character NMT, as attention won't be able to attend individual characters.

\section{Conclusion and Future Work}
\label{conclusion}
In this work, we have tried to analyze the scenarios where SMT performs better than NMT and vice-versa. Also, we have tried to find out whether MT models give better outputs when trained with simple sentences rather than when trained using sentences of various complexities. 

As a future prospect, we would like to take the "other" (Complex+Compound) sentence pairs and simplify it, so that the whole MT models can be trained using more simple sentences. Also, we would like to increase the number of LSTM encoding and decoding layers as well as include embeddings like ConceptNet\footnote{https://github.com/commonsense/conceptnet-numberbatch} in our future works.

\section*{Acknowledgments}
This work is supported by Media Lab Asia, MeitY, Government of India, under the Visvesvaraya PhD Scheme for Electronics \& IT.

\bibliography{acl2018}
\bibliographystyle{acl_natbib}

\end{document}